\documentclass[runningheads]{llncs}
\usepackage{graphicx}

\usepackage{amsmath,amssymb,amsfonts}
\usepackage[numbers]{natbib}
\usepackage{algorithm}      
\usepackage{algpseudocode}  
\usepackage{subfig}
\DeclareMathOperator\erfc{erfc}

\begin{document}
\title{The Randomness of Input Data Spaces is an\\ A Priori Predictor for Generalization\\}
\titlerunning{The Randomness of Input Data Spaces is a Predictor for Generalization}
%




 \author{Martin Briesch\inst{1}\orcidID{0000-0002-8209-1465} \and
 Dominik Sobania\inst{1}\orcidID{0000-0001-8873-7143} \and
 Franz Rothlauf\inst{1}\orcidID{0000-0003-3376-427X}}
 \authorrunning{M. Briesch, et al.}
\institute{Johannes Gutenberg-University, Mainz, Germany 
\email{\{briesch,dsobania,rothlauf\}@uni-mainz.de}}
\maketitle              
\begin{abstract}
    Over-parameterized models can perfectly learn various types of data distributions, however, generalization error is usually lower for real data in comparison to artificial data. This suggests that the properties of data distributions have an impact on generalization capability. This work focuses on the search space defined by the input data and assumes that the correlation between labels of neighboring input values influences generalization. If correlation is low, the randomness of the input data space is high leading to high generalization error. We suggest to measure the randomness of an input data space using Maurer's universal. Results for synthetic classification tasks and common image classification benchmarks (MNIST, CIFAR10, and Microsoft's cats vs. dogs data set) find a high correlation between the randomness of input data spaces and the generalization error of deep neural networks for binary classification problems.
\end{abstract}
\section{Introduction}
\label{intro}

While deep neural networks (DNN) have gained much attention in many machine learning tasks \citep{lecun2015deep}, there is still only limited theory explaining the success of DNN. Especially the generalization abilities of DNNs have challenged classical learning theory as standard approaches like VC-dimension \citep{vapnik2013nature}, Rademacher complexity \citep{bartlett2002rademacher}, or uniform stability \citep{bousquet2002stability} fail to explain the generalization behavior of  over-parameterized DNNs \citep{zhang2016understanding}. Most of the existing theory approaches look at the hypothesis space of the model and the properties of the learning algorithm; properties of the data distribution (as well as the machine learning tasks) are addressed to a much lower extent.

Focusing on the data distribution, \citep{zhang2016understanding} observed a lower generalization capability of DNNs when randomizing natural data.  Arpit~et.~al.~\citep{arpit2017closer} find that learning on real data behaves differently than learning on randomized data. DNNs seem to work content-aware and learn certain data points first.
Thus, there is evidence that the properties of the input data distribution have an influence on the generalization capabilities of DNNs and natural data has properties that enable DNNs to perform well. This raises the question why DNNs perform well on supervised learning tasks with natural data signals.

This paper studies how the properties of training data influence the generalization capability of DNNs. We assume a label landscape $(X, f, \mathcal{N})$ with the set of training data $X$, the labeling function  $f:X\rightarrow Y$ that assigns a label $y\in Y$ to each training instance $x\in X$, and a neighborhood mapping  $\mathcal{N}:X\rightarrow 2^X$ which assigns to each input $x$ a set of neighboring inputs. We suggest that the properties of the label landscape formed by the training data influences the generalization behavior of DNNs. 

To measure the properties of the training data, we perform a random walk through the label landscape $(X, f , \mathcal{N})$. A random walk with $N$ steps iteratively selects a neighboring training instance $x_i$ (based on a distance metric) and returns the corresponding label $y_i$. Thus, it creates a sequence of labels $y^N$. We expect that the randomness of $y^N$ (for example measured by Maurer's universal test) influences the generalization capability of DNNs. If Maurer's universal test indicates that $y^N$ is a random sequence, then generalization is expected to be low; in contrast, if $y^N$ is non-random  (which means the per-bit entropy of the sequence is low), DNNs are expected to be able to learn well and show high generalization capability for this particular data distribution. Thus, we suggest that the randomness of a sequence of binary labels generated by a random walk through the input data space is a good predictor for the expected generalization capability of DNNs. 

We present evidence and experimental results for four types of problems. First, we follow the approach suggested by \citep{zhang2016understanding} and systematically randomize the labeling function  $f:X\rightarrow Y$ by assigning the label \(y\) independently at random with probability \(v\). With stronger randomization of the labels, the resulting sequence $y^N$ created by a random walk has higher randomness according to Maurer's universal test and generalization decreases. We present results for different binary instances of synthetic test problems where we know the decision boundaries (an XOR type problem, a majority vote problem, and a parity function problem). 
Second, we study binary instances of MNIST \citep{lecun1998gradient} and CIFAR10 \citep{krizhevsky2009learning} using the same randomization method as in the previous experiments and extend the results with experiments where we randomize the training instances $x\in X$. For the extension, we consider four different variants. We either perform a random permutation $\pi:x\rightarrow x$ of all input variables of the training data (\emph{PermutGlobal}), perform a random permutation of all variables for all training instances (\emph{PermutInd}), draw each input value randomly from a Gaussian distribution matching the original distribution of the input values (\emph{GaussianInd}), or draw each input value from a white noise distribution (\emph{NoiseInd}). The results indicate that Maurer's universal test applied to the sequence $y^N$ is a good predictor for the expected generalization capability of a DNN.
Finally, we focus on binary instances of the more complex cats vs.~dogs data set \citep{elson2007asirra} and distinguish between training instances that are either easy or difficult to learn by a DNN. Experimental results confirm that the randomness of $y^N$ is a good indicator of the expected generalization. 

In Sect.~\ref{theory}, we describe preliminaries and present Maurer's universal as a novel measure for the randomness of data sets and related supervised learning tasks. Sect.~\ref{experiments} describes the experimental setting and presents the results. In Sect.~\ref{relWork}, we 
give an overview of related work before concluding the paper in Sect.~\ref{conclusion}. Sect.~\ref{limits} describes the limitations and future research directions.

\section{Randomness of Data Spaces}
\label{theory}

Consider a data set \(\mathcal{D}\) consisting of a finite number \(m\) of pairs \((x_i, y_i)\) where \(x \in X\) and \(y \in Y\). 
$x_{ij}$ denotes the value of the $j$-th input variable of the vector $x_i$; $y_i$ denotes the corresponding label. All pairs are drawn i.i.d. from the population distribution \(P_{XY}\). The goal of a machine learning model in a supervised classification task is to find a function \(h^*\) from a hypothesis space \(\mathcal{H}\) given a loss function \(l\) that minimizes the population risk \(R(h)\):

\[R(h) = \mathbb{E}[l(h(X),Y)]\] 
\[h^*= \arg \min_{h\in \mathcal{H}} R(h) \]

Usually, the model does not have access to the complete distribution \(P_{XY}\) but rather only to the data set \(\mathcal{D}\). Therefore, a common approach in machine learning is to minimize the empirical risk \(R_{emp}(h)\) on the given data \(\mathcal{D}\):

\[R_{emp}(h) = \frac{1}{m}\sum_{i=1}^{m}l(h(x_i),y_i)   \] 
\[\hat{h}= \arg \min_{h\in \mathcal{H}} R_{emp}(h) \]

Unfortunately, the empirical risk can be significantly different from the population risk. This makes bounding the gap between \(R(h)\) and \(R_{emp}(h)\), also called \emph{generalization}, a central challenge in machine learning \citep{vapnik1992principles}.

In theory, given a sufficient amount of parameters and training time, a multilayer neural network can approximate any function \(h\) arbitrarily well  \citep{cybenko1989approximation, hornik1991approximation}. Thus, any data set \(\mathcal{D}\) can be learned by a large enough model. This is confirmed by empirical studies where complex DNN models can fit both data from natural signals as well as random data \citep{zhang2016understanding}. Learning arbitrary $h$ can be achieved by standard DNN models without changing any hyperparameters, neither for the model nor for the used learning algorithm. When fitting DNN models to either natural signals or random data, \citep{zhang2016understanding} as well as  \citep{arpit2017closer} observed differences in the generalization error. For natural signals, usually the generalization error is low; for random or randomized data, generalization error is high. 

We believe that the differences in generalization error $g_{\mathrm{err}}$ between different data sets can be explained by the properties of the label landscape defined on the data set $\mathcal{D}$. Analogously to fitness landscapes known in other domains, we define a label landscape $(X,f,\mathcal{N})$, where the labeling function $f:X\rightarrow Y$ assigns a label $y\in Y$ to each training instance $x\in X$ and a neighborhood mapping $\mathcal{N}:X\rightarrow 2^X$ assigns to each input $x\in X$ a set of neighboring inputs. The labeling function $f$ is defined by the input data; the neighborhood mapping $\mathcal{N}$ is usually problem-specific and defines which input/training data is similar to each other \citep{wright1932roles, herrmann2016fitness}. Instead of defining $\mathcal{N}$ on the  raw input data, we can also define $\mathcal{N}$ on an underlying manifold representing the data. 

Using a label landscape defined on the input data, we can calculate relevant properties like the correlation between neighboring data points. Such measures are relevant for combinatorial optimization problems as problems, where the objective values of neighboring solutions are uncorrelated, are difficult to solve \citep{Jones:95,rothlauf2011design}. If fitness values (labels) of neighbors in the input space are uncorrelated, the no free lunch theorem holds \citep{wolpert1995no, wolpert1997no, wolpert1996lack, wolpert1996existence} and optimization methods can not beat random search. The situation is similar for non-parametric machine learning methods like kernel machines which rely on the smoothness prior \(h(x)\approx~h(x+\epsilon)\). The smoothness prior assumes that the properties of neighboring inputs (either measured in time or in space) are similar and do not abruptly change. Consequently, kernel machines have problems to learn non-local functions with low smoothness  \citep{bengio2006curse}, although deep learning is able to learn some variants of non-local functions  \citep{imaizumi2019deep}. 

\begin{algorithm}[!ht]
	\caption{Random walk}
	\label{alg:randomwalk}
	\begin{algorithmic}[1]
	    \State Select random start point $x_{0} $
	    \State Initialize \(y^N = [y_{0}]\)
		\For {$z=1,2,\ldots,N$}
		    \State Select \( x_{z}\) randomly from the neighborhood \(\mathcal{N} (x_{z-1}) \)
		    \State Append \( y_{z}\) to \( y^N \)
		\EndFor
	\end{algorithmic}
\end{algorithm}

We suggest to capture the correlation between labels of neighboring input values (taken from the given data set $\mathcal{D}$) by performing a random walk through $(X, f,\mathcal{N})$ and analyzing the resulting sequence $y^N$ of labels.
Algorithm~\ref{alg:randomwalk} shows the random walk as pseudo-code. 
We initialize $y^N$ with the label $y_0$ of a random start point $x_0$ (lines 1-2) and perform $N$ times a step of the random walk appending the label $y_z$ of a randomly selected $x_z$ from the neighborhood \(\mathcal{N} (x_{z-1}) \) (lines 3-6).

\begin{figure*}[t]
  \centering
  \subfloat[Local pattern.]{\includegraphics[width=0.25\textwidth]{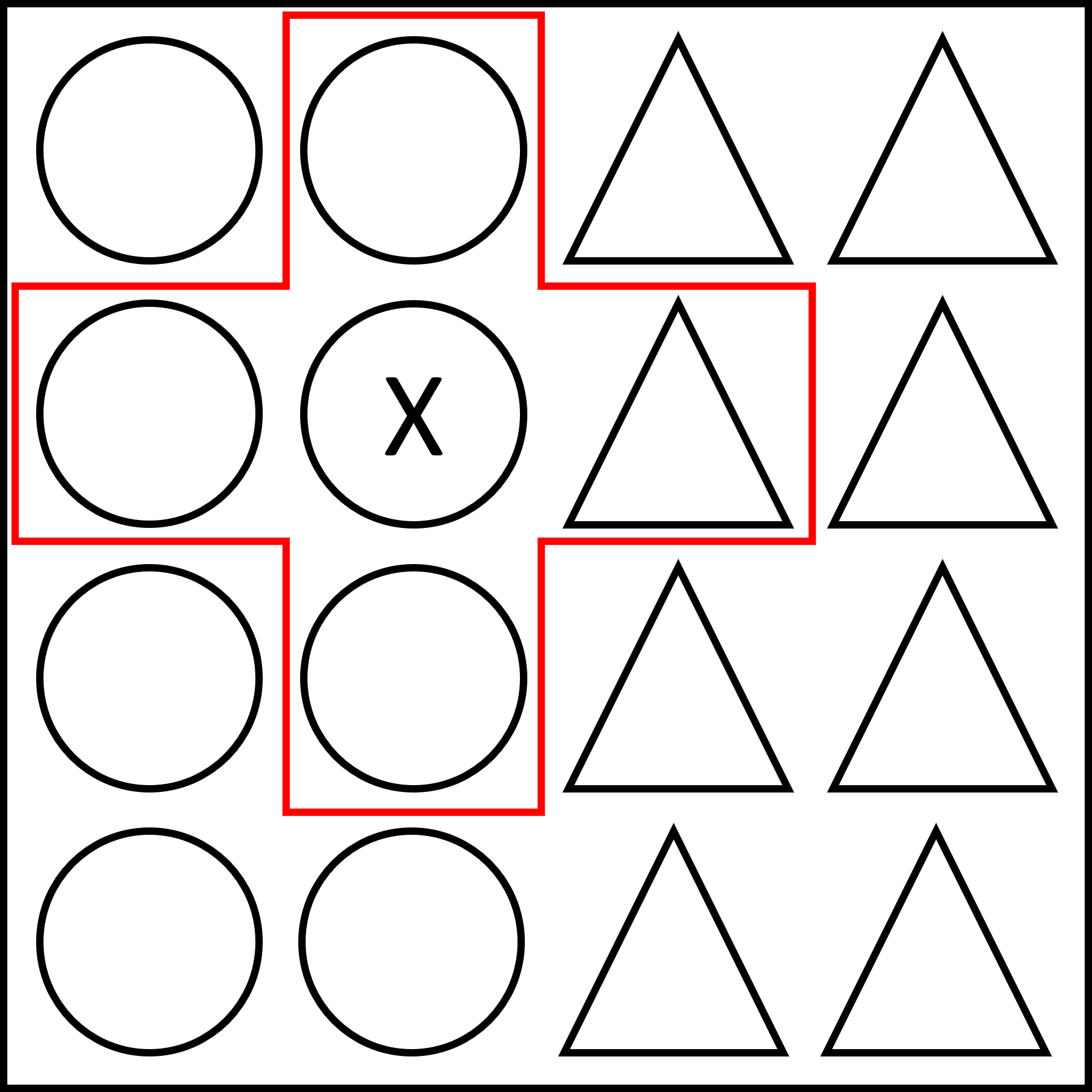}\label{fig:local}}
  \;\;\;
  \subfloat[Non-local pattern.]{\includegraphics[width=0.25\textwidth]{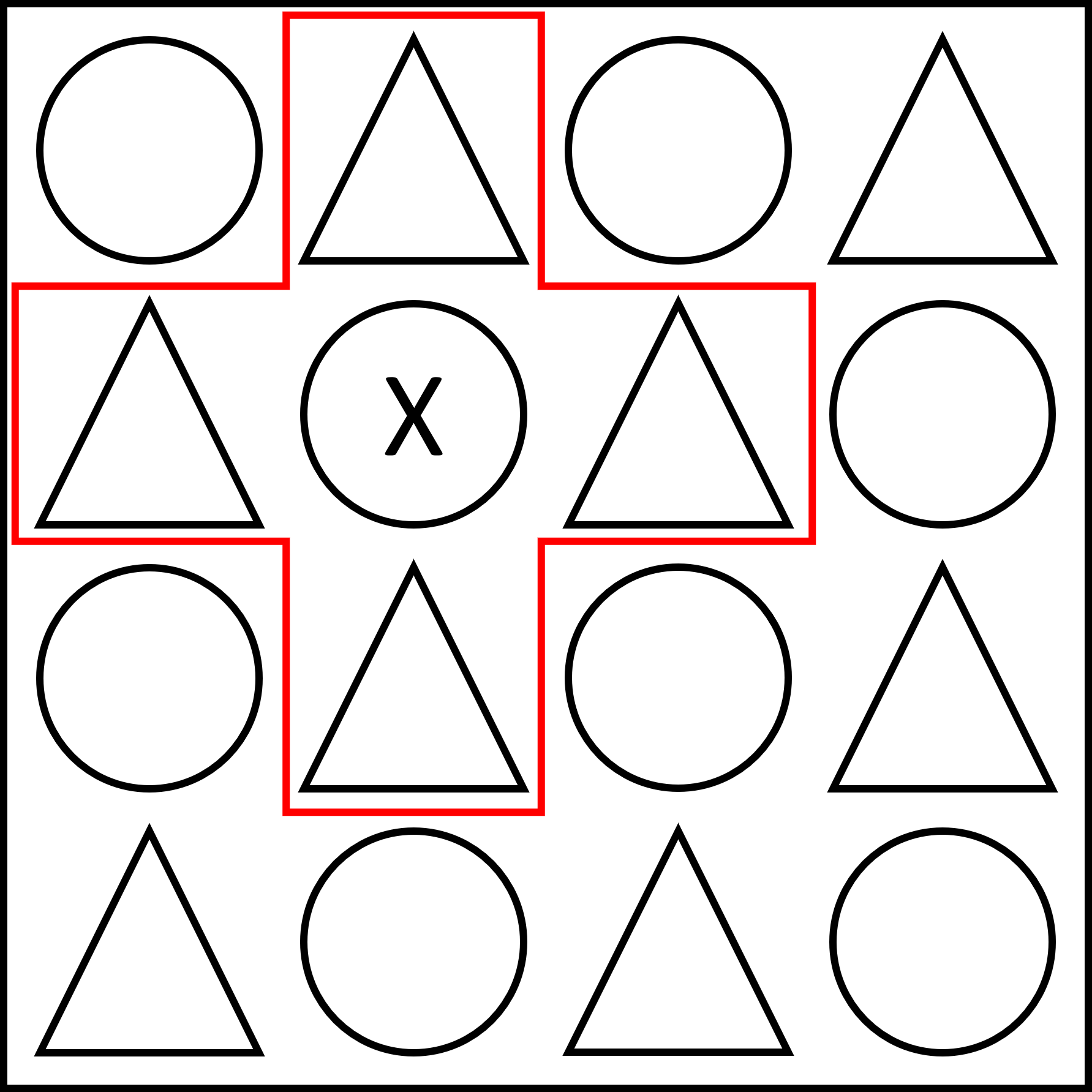}\label{fig:nonlocal}}
    \;\;\;
  \subfloat[Random pattern.]{\includegraphics[width=0.25\textwidth]{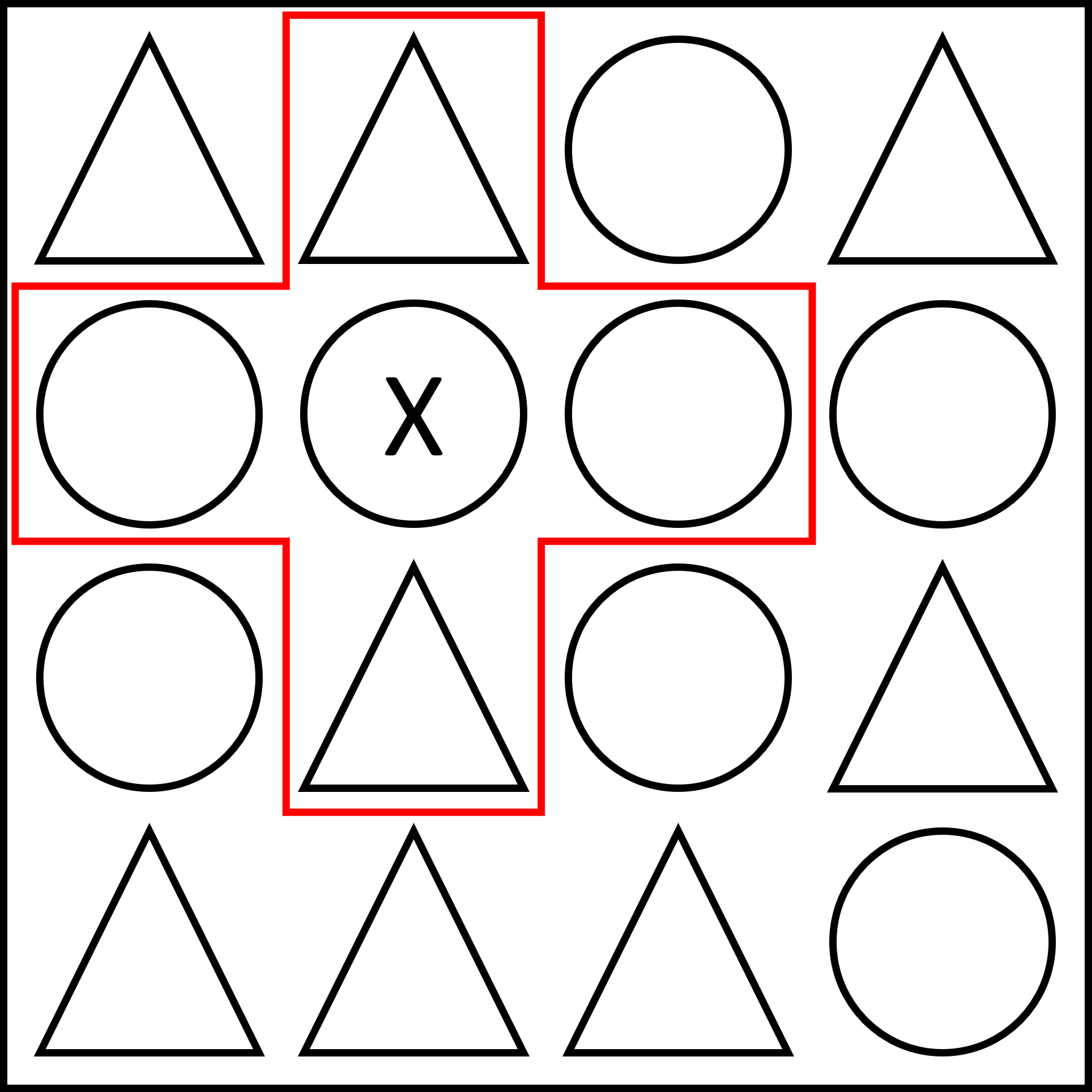}\label{fig:random_pattern}}

  \caption{Resulting landscapes for different example binary classification problems. Each input data has four neighbors. 
  (a) easy problem with high correlation between labels of neighboring input data 
  (b) non-local, but easy problem with low randomness in $y^N$
  (c) non-local and difficult problem, where each input data has a randomly chosen label.}
  \label{fig:neighbourhood-toyexample}
\end{figure*}

We expect that the randomness of $y^N$ influences the generalization ability of DNNs trying to learn the properties of $\mathcal{D}$. For example, we assume a binary classification problem that can easily be learned and linearly separated (see Figure~\ref{fig:local}). When performing a random walk through the space of input values, the value of the corresponding label $y_i$ rarely changes and the randomness of the resulting sequence $y^N$ is low. Situation is different, if we assign random labels to the input data points (Fig.~\ref{fig:random_pattern}). Then, the resulting sequence $y^N$ is random. In contrast, Fig.~\ref{fig:nonlocal} shows the landscape of the parity problem, which can be well learned using DNN \citep{imaizumi2019deep} but is a non-local problem. When performing a random walk through such a landscape, the resulting sequence $y^N$ is non-random but highly structured as the labels of neighboring input data points are always different. This  property of the classification problem  can be learned by an appropriate model.

To measure the statistical randomness of a binary sequence $y^N$, we suggest using Maurer's universal test \(T_U\) \citep{maurer1992universal, coron1998accurate}. The purpose of Maurer's universal test is to measure the entropy in the sequence $y^N$. Other possibilities to measure the statistical randomness of a sequence are the Wald–Wolfowitz runs test \citep{wald1940test}, which measures the number of label changes, or autocorrelation tests \citep{box1976time}. We choose Maurer's universal test as it is able to detect also high-order as well as non-linear dependencies in a sequence.  

We use the statistical test Maurer's universal \(T_U\) to test if the source process of the sequence is random \citep{maurer1992universal, coron1998accurate}. Maurer's universal takes the sequence \(y^N\) of binary labels \(y\) (from \(B = \{0,1\}\)) as input. The test has three parameters \(\{L,Q,K\}\). It partitions the sequence in blocks of length \(L\) with \(Q\) blocks used for initializing the test and \(K\) blocks to perform the test. Thus,  $N = (Q+K)L$ and \(b_n(y^N) = [y_{L(n-1)+1}, \ldots, y_{Ln}] \). The test function \(f_{T_U}: B^N \rightarrow \mathbb{R}\) measures the per-bit entropy and is defined as 

\[f_{T_U}(y^N) = \frac{1}{K} \sum_{n=Q+1}^{Q+K}\log_2 A_n(y^N),\]

where

{ \[A_n(y^N)=\left\{ \begin{array}{l} 
                    n \mbox{ , if } \forall a < n, b_{n-a}(y^N) \neq b_n(y^N) \\ 
                    \min\{a:a \geq 1, b_n(y^N) = b_{n-1}(y^N)\} \mbox{ , otherwise.} \\
                    \end{array}\right.
                    \]
 }

This test function can be used to compute the \(p\in[0,1]\) value

\[ p = \erfc\left( \left\vert \frac{f_{T_U}-\text{expectedValue}(L)}{\sqrt{2} \sigma}  \right\vert \right),
\]

where \({\erfc}\) is the complementary error function. \(\text{expectedValue}(L)\) and \(\sigma\) are precomputed values \citep{maurer1992universal}. The \(p\) value measures the confidence whether the process is non-random. Thus, low values of \(p\) indicate a high probability that the  process is non-random.

If Maurer's universal test indicates that $y^N$ is a random sequence (high values of $p$), then the generalization capability of a DNN applied to this data set $\mathcal{D}$ is expected to be low; in contrast, if $y^N$ is non-random  (which means the per-bit entropy of the sequence is low), DNNs are expected to be able to learn well the structure of $\mathcal{D}$ and show high generalization capability. Thus, we suggest that the randomness of a sequence of binary labels generated by a random walk through the input data space is a good predictor for the expected generalization capability of DNNs learning the input data.

\section{Experiments and Discussion}
\label{experiments}

To study how the properties of input data influences the generalization capability of DNNs, we randomize all studied data sets to different degrees as suggested by \citep{zhang2016understanding} and perform random walks through the label landscapes $(X, f, \mathcal{N})$ as described in Algorithm \ref{alg:randomwalk}. For all considered data sets, we perform $30$ random walks with $N=1,000,000$ steps and calculate the 
confidence $p$ for the resulting sequence $y^N$ of labels. As data sets, we use synthetic classification tasks as well as on the common classification benchmarks MNIST \citep{lecun1998gradient}, CIFAR10 \citep{krizhevsky2009learning}, and the cats vs.~dogs data set \citep{elson2007asirra}. For each test problem, the input data is split into 80\% train and 20\% test data.

For the synthetic classification tasks as well as MNIST, we train a multilayer perceptron (MLP) consisting of two hidden layers with 4,096 neurons each and ReLU activation functions. For CIFAR10 and the cats vs.~dogs data set, we use a small convolutional network (CNN) with three convolutional layers with 32/64/64 filters of kernel size 3x3 followed by a dense layer with 256 hidden neurons. After each convolutional layer we use 2x2 MaxPooling and all layers use ReLU activation functions. The models are trained with the Adam optimizer \citep{kingma2014adam} until convergence to 100\% accuracy on the train data. Thus, test error is identical to the generalization error $g_{\mathrm{err}}$. 

All experiments were conducted on a workstation using an AMD Ryzen Threadripper 3990X 64x2.90GHz, an NVIDIA GeForce TITAN RTX and 128GB DDR4 RAM. The DNNs were implemented using Tensorflow 2 \citep{abadi2016tensorflow}. 

\subsection{Synthetic Classification Problems with Known Decision Boundaries}
\label{binary}

\begin{figure*}[!t]
  \centering
  \includegraphics [width=0.24\textwidth]{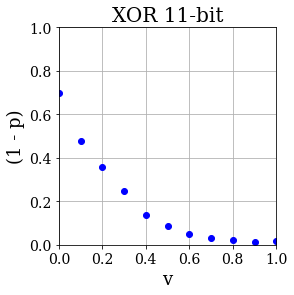}
  \includegraphics [width=0.24\textwidth]{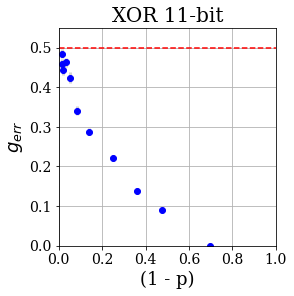}
  \includegraphics [width=0.24\textwidth]{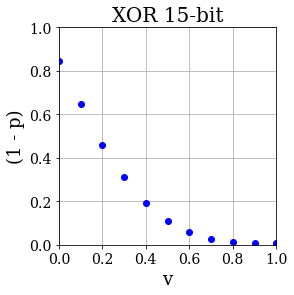}
  \includegraphics [width=0.24\textwidth]{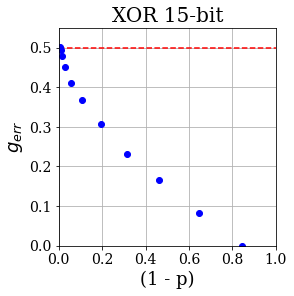}
  \includegraphics [width=0.24\textwidth]{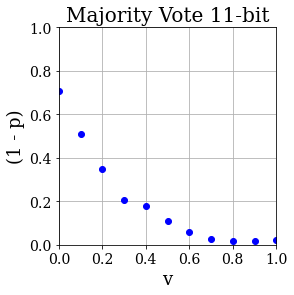}
  \includegraphics [width=0.24\textwidth]{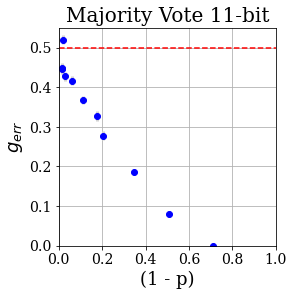}
  \includegraphics [width=0.24\textwidth]{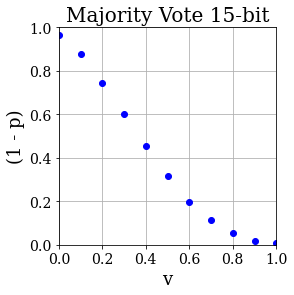}
  \includegraphics [width=0.24\textwidth]{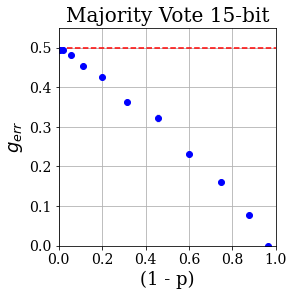}
  \includegraphics [width=0.24\textwidth]{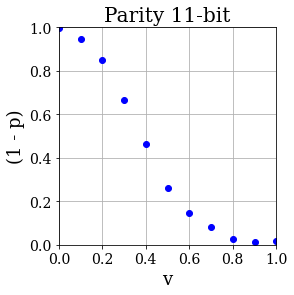}
  \includegraphics [width=0.24\textwidth]{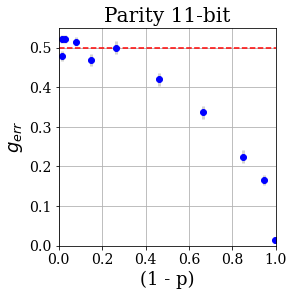}
  \includegraphics [width=0.24\textwidth]{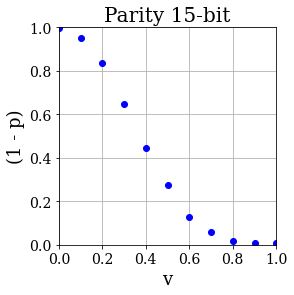}
  \includegraphics [width=0.24\textwidth]{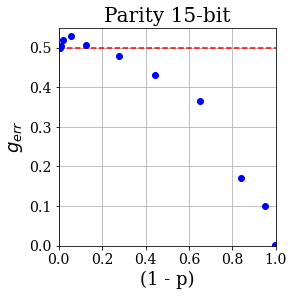}
  \caption{$(1-p)$ over the randomization level $v$ and generalization error $g_{\mathrm{err}}$ over $(1-p)$ for all studied synthetic 
  classification problems (XOR, majority vote, and parity) for $d=11$ and $d=15$. The dashed line indicates performance of random guessing. All results are averaged over $30$ runs.}
  \label{fig:bit-problems}
\end{figure*}

To analyze whether the suggested measure $p$ properly captures the randomness of a problem for both, local and non-local patterns, 
we first study problems where we already know the the classification problem's decision boundaries.  
We select three synthetic $d$-bit binary classification problems. The first one is a XOR type problem with binary input vectors \( x_i \) (\( x_{ij} \in \{0,1\}\)). The label of each vector $x_i$ depends on the first two input variables while the remaining features hold no explanatory power:
\[y_i =     \left\{ \begin{array}{rcl} 
                    1 & \mbox{for}& x_{i,1} = x_{i,2} \\ 
                    0  & \mbox{for} & x_{i,1} \neq x_{i,2} \\
                    \end{array}\right.
\]
The second test problem uses the same binary input vectors \( x_i \). The label is determined by the majority vote over the elements \(x_{i,j}\):
\[y_i =     \left\{ \begin{array}{rcl} 
                    1 & \mbox{for}& \sum_{j=1}^{d}(x_{i,j}) \geq \frac{d+1}{2} \\ 
                    0  & \mbox{for} & \sum_{j=1}^{d}(x_{i,j}) < \frac{d+1}{2} \\
                    \end{array}\right.
\]
The third test problem also uses binary input vectors \( x_i \). The label of each vector is determined by the parity function:
\[y_i =     \left\{ \begin{array}{rllc} 
                    1 & \mbox{if}& \sum_{j=1}^{d}(x_{i,j}) &\mbox{is even} \\ 
                    0  & \mbox{otherwise}  \\
                    \end{array}\right.
\]

For all synthetic classification tasks, we study instances of different size $d\in\{11,15\}$ and corrupt the labeling processes by changing 
each label \(y\) with probability \(v\) to a random class in the training and test set (see \citep{zhang2016understanding}) 
to construct different instances of the tasks with varying degrees of structure.
The used data set $\mathcal{D}$ consists of all possible input vectors, as we assume that \(X = \mathcal{D}\).
The neighborhood function \(\mathcal{N}(x)\) maps each input $x_i \in X$ to a set of inputs $x \in X$ that are different from \(x_i\) in one position \(x_{i,j}\). We measure the randomness of \(y^N\) (constructed by the random walk) using Maurer's universal and compare it to the generalization performance of the MLP/CNN.

Figure~\ref{fig:bit-problems} plots the measure $(1-p)$ over the randomization level $v$ 
and the generalization error $g_{\mathrm{err}}$ over $(1-p)$ for all studied synthetic classification problems for $d=11$ and $d=15$. For comparison, the dashed line indicates the performance 
of random guessing. All results are averaged over $30$ runs. 

We expect that for higher values of $v$ (which leads to a higher randomness of $y^N$ and a lower correlation between neighboring inputs) the inherent structure of the classification problem sets declines which leads to lower generalization. 
The results confirm this expectation, as we can observe lower values of $(1-p)$ for larger values of $v$ as well as a lower generalization 
error $g_{\mathrm{err}}$ 
for high values of $(1-p)$. For the considered test problems, the measure $(1-p)$ is a good predictor for generalization as Pearson's $r$ correlation coefficient between generalization error $g_{\mathrm{err}}$ and $(1-p)$ is lower than $-0.94$ for all studied problem instances. 
This holds not only for small problems ($d=11$) but also for larger problem instances ($d=15$). Furthermore and contrary to the smoothness prior, the measure $(1-p)$ correctly detects structure (non-randomness) not only in local (XOR, majority vote) but also in non-local (parity) patterns.

\subsection{Natural Data with Unknown Decision Boundaries}
\label{mnist_binary}

To verify whether our findings also hold on natural data, we extend our experiments to the MNIST and CIFAR10 data sets. We consider a binary classification version of those problems and (as before) corrupt the labeling function $f$ by randomizing each label \(y\) with probability \(v\). Again, we study the randomness of \(y^N\) (created by a random walk) and compare it to 
the generalization capability of MLP/CNN. However, since the true decision variables for the MNIST and CIFAR10 problems do not lie in the raw input matrix but rather
are represented by latent variables in an underlying manifold \citep{goodfellow2016deep}, we first approximate such manifold by reducing the 
dimension of the input data with a variational autoencoder \citep{hinton2006reducing, kingma2013auto}. Consequently, we define the neighborhood
\(\mathcal{N}(x)\) on \(\mathcal{D}\) as the set of \(k\) nearest data points measured by Euclidean distance inside this manifold. In our experiments, we chose \(k = 10\). 

\begin{figure*}[t]
  \centering
  \includegraphics [width=0.98\textwidth]{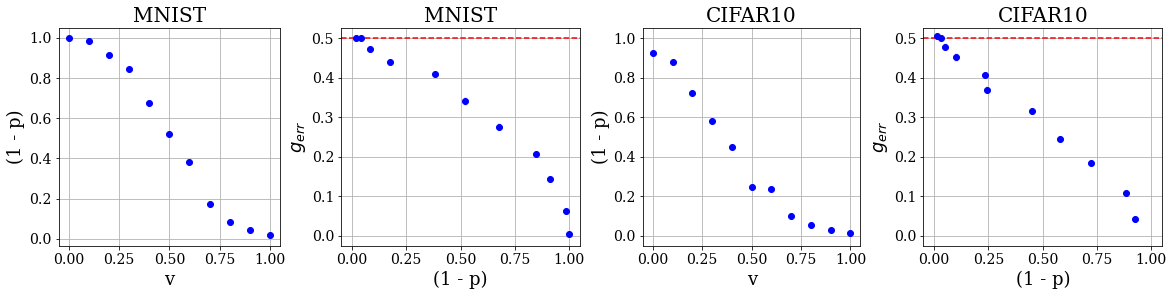}
  \caption{$(1-p)$ over the randomization level $v$ and the generalization error $g_{\mathrm{err}}$ over $(1-p)$ for the binary versions of MNIST and CIFAR10.}
  \label{fig:2class_randY}
\end{figure*}

Figure \ref{fig:2class_randY} plots the measure $(1-p)$ over the randomization probability $v$ 
and the generalization error $g_{\mathrm{err}}$ over $(1-p)$ for the binary versions of MNIST and CIFAR10. The dashed line indicates the generalization error $g_{\mathrm{err}}$ of random guessing. Again, all results are averaged over $30$ runs. 

\begin{figure*}[t]
  \centering
  \includegraphics [width=0.98\textwidth]{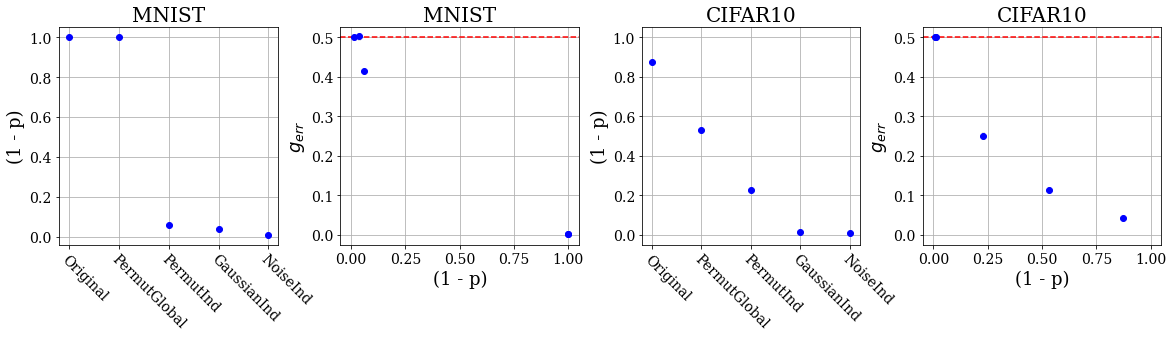}
  \caption{$(1-p)$ over four variants of randomization (\emph{PermutGlobal}, \emph{PermutInd}, \emph{GaussianInd} and \emph{NoiseInd})
and generalization error $g_{\mathrm{err}}$ over $(1-p)$ for the binary versions of MNIST and CIFAR10. 
}
  \label{fig:2class_randX}
\end{figure*}

As expected, we also find a strong correlation between $p$ and $g_{\mathrm{err}}$ for natural signals. Again, we observe lower values of  $(1-p)$ for larger values of $v$ and a lower generalization error $g_{\mathrm{err}}$ for high values of $(1-p)$. The Pearson's $r$ correlation coefficient between 
generalization error $g_{\mathrm{err}}$ and $(1-p)$ is lower than $-0.97$ for both problem sets
indicating that $(1-p)$ is a good approximation of the expected generalization error also on natural data.

To study the effects of different types of randomization of $f$, we now permutate the inputs $x \in X$ instead of the labels $y \in Y$. We consider four different variants: 1) a random permutation $\pi:x\rightarrow x$ of all input variables $x_{ij}$ of the training data (denoted as \emph{PermutGlobal}), 2) a random permutation of all variables for all training instances (\emph{PermutInd}), 3) replacing a variable value by a random input value from a Gaussian distribution matching the original distribution of input values (\emph{GaussianInd}), and 4) replacing a variable value by a value randomly drawn from a white noise distribution (\emph{NoiseInd}). As before, we study whether the randomness of $y^N$ is related to the generalization error.

Figure \ref{fig:2class_randX} plots $(1-p)$ over the four different variants of randomization  
and the resulting generalization error $g_{\mathrm{err}}$ over $(1-p)$. Again, the dashed line indicates the performance of random guessing. All results are averaged over $30$ runs.

Again, we find a strong correlation (Pearson coefficient $<-0.99$) between generalization error $g_{\mathrm{err}}$ and $(1-p)$. For \emph{PermutGlobal}, we observe a lower effect of randomization for MNIST in comparison to CIFAR10 as the neighborhood of the input data space is more relevant for CIFAR10 than MNIST. For MNIST, the value of a pixel $x_{ij}$ also has a meaning independently of its neighboring pixels (e.g.~some pixels are always activated for a specific label). In contrast for CIFAR10, destroying the neighborhood of a pixel $x_{ij}$ by placing it next to other, randomly selected pixels makes it much more difficult for the DNN to build a meaningful model. As a result,  $(1-p)$ is lower for CIFAR10. 
For \emph{PermutInd}, results are different as the only signal that is left after randomization is the difference in mean and standard deviation of input variables. The differences are higher in CIFAR10 training instances which  makes the problem more structured (leading to a lower generalization error) in comparison to MNIST. Both cases are properly captured by $(1-p)$.

\subsection{Studying Randomness of Input Data Spaces without Randomization}
\label{cats_vs_dogs}

\begin{figure}[!t]
    \centering
    \includegraphics [width=0.33\linewidth]{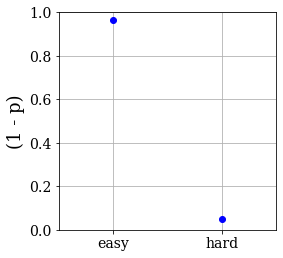}
    \includegraphics [width=0.33\linewidth]{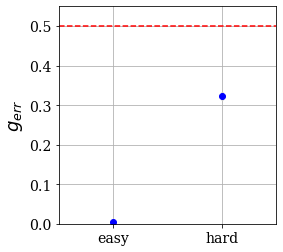}
    \caption{$(1-p)$ for the easy and hard data samples with the corresponding generalization error $g_{\mathrm{err}}$.}
    \label{fig:cats_vs_dogs}
\end{figure}

While our previous experiments studied the relationship between the randomness of input data spaces measured by $(1-p)$ and generalization error for different degrees and variants of randomization, we now investigate differences in the randomness of input data spaces for easy versus hard data samples. Thus, we do not randomize neither $f$ (Sect.~\ref{binary}) nor $X$ 
(Sect.~\ref{mnist_binary}), but create data samples with different properties from $\mathcal{D}$ following an approach suggested by \citep{arpit2017closer}. Consequently, we first train 100 CNNs for 1 epoch on a large data set (cats vs.~dogs). Then, we select two subsets (easy versus hard) from $\mathcal{D}$ by selecting the on average 10,000 best and 10,000 worst classified examples for the easy and hard subset, respectively. We expect that Maurer's universal is a good indicator for the differences in randomness of these samples and the resulting generalization error $g_{\mathrm{err}}$.

Figure \ref{fig:cats_vs_dogs} plots $(1-p)$ for the easy and hard data samples as well as the corresponding generalization error $g_{\mathrm{err}}$. The dashed line indicates the performance of random guessing. Results are averaged over $30$ runs.
We find that a high value of $(1-p)$ (indicating a high randomness in  $y^N$) correspond to a low generalization error $g_{\mathrm{err}}$ on the easy sample and vice versa on the hard sample confirming the prediction quality of Maurer's universal. For the easy sample, the generalization error is almost zero which corresponds to a high value of $(1-p)\approx 1$ indicating a low randomness of $y^N$ and a high structure of the classification problem. Thus the easy data set can be learned by a DNN model with low generalization error. For the hard sample, the randomness of $y^N$ is high indicating a low correlation between the labels of neighboring training points.  

\section{Related Work}
\label{relWork}

Bounding the best and worst case for generalization error is a key challenge in machine learning. Traditional learning theory provides such bounds either from a complexity point of view \citep{vapnik2013nature,bartlett2002rademacher} or using a stability based approach \citep{bousquet2002stability}. However, studies suggest that these generalization bounds might not be sufficient to capture the generalization problem, especially in an over-parameterized setting \citep{zhang2016understanding, NEURIPS2019_05e97c20, belkin2019reconciling}.
This leads to work on extending and sharpening the traditional bounds for neural networks by introducing norms \citep{NIPS2017_b22b257a, kawaguchi2017generalization, neyshabur2015norm,NIPS2017_10ce03a1,neyshabur2018role, golowich2018size, liang2019fisher} or using  PAC-Bayes approaches \citep{neyshabur2018pac, dziugaite2017computing, zhou2018nonvacuous, arora2018stronger}. A different direction of research studies the implicit regularization from gradient descent methods to explain generalization \citep{hardt2016train, soudry2018implicit, l.2018a, NEURIPS2019_c0c783b5}. 

However, most of these approaches depend on posterior properties of a trained neural network. In contrast,
\citep{arpit2017closer} find that the data itself plays an important role in generalization. 
Therefore, other work focuses on the properties of data in context of generalization.
Ma~et.~al.~\citep{ma2018priori} provide a prior estimate using properties of the true target function and
\citep{arora2019fine} derive a data-depended complexity measure using the Gram matrix of the data 
and \citep{farnia2020fourier} analyze the properties of classification problems using Fourier analysis. 
The method suggested in this paper differs as we take a label landscape perspective to derive a generalization estimate.

\section{Conclusion}
\label{conclusion}

This paper introduced a landscape perspective on data distributions in order to explain generalization performance of DNNs. 
We argued that the input data defines a label landscape and the correlation between labels of neighboring (similar) input values influences generalization. 
We measure the correlation of the labels of neighboring input values by performing a random walk through the input data space and use Maurer's universal to measure the randomness of the resulting label sequence $y^N$. A more random sequence indicates a less learnable structure in the data leading to poor generalization. At the extreme, if there is no correlation between the labels of neighboring inputs, generalization error is maximal. We performed experiments for a variety of problems to validate our hypothesis and found that the randomness (measured by Maurer's universal) of the label sequence $y^N$ indeed can serve as an a priori indicator of the expected generalization error for a given data set. We presented results for both synthetic problems as well as real world data sets and found a high correlation between the randomness of the label sequence $y^N$ and the generalization error. We conclude that a label landscape view on the data provides valuable insight into the generalization  capability of DNN.

\section{Limitations and Future Work}
\label{limits}

Our approach provides insights and an a priori indicator for generalization in a binary classification case. However, there are a few limitations due to the use of Maurer's universal test. As the test is only designed for a binary source processes, it is not applicable to multi-class problems. Therefore, in future work we will study randomness measures for integer sequences. 

If the decision variables are not known, our method depends on the approximation of the underlying manifold, for which we assume an Euclidean space. Approximating such a manifold can be challenging for more difficult data sets. Studying the impact of this approximation and different distance measures for the neighborhood could lead to a better understanding of our findings.
%
%
%
\bibliographystyle{splncs04}
\bibliography{bibfile}

\end{document}